\documentclass{article}
\usepackage{spconf,amsmath,graphicx,cleveref,caption,subcaption,dsfont,multirow,multicol,color,xcolor}

\usepackage[pscoord]{eso-pic}

    \def\1{{\bf
1}}   

\newcommand{\placetextbox}[3]{
  \setbox0=\hbox{#3}
  \AddToShipoutPictureBG*{
    \put(\LenToUnit{#1\paperwidth},\LenToUnit{#2\paperheight}){\vtop{{\null}\parbox{\textwidth}{\centering #3}}}%
  }%
}%

\title{WEIGHT REPARAMETRIZATION FOR BUDGET-AWARE NETWORK PRUNING}
\name{Robin Dupont$^{\star \dagger}$ \qquad Hichem Sahbi$^{\star}$ \qquad Guillaume Michel$^{\dagger}$}
  
  \address{$^{\star}$ Sorbonne Université, LIP6, Paris \\
      $^{\dagger}$ Netatmo, Boulogne-Billancourt}

\begin{document}

\placetextbox{0.085}{0.96}{\color{gray}\small\noindent © 2021 IEEE. Personal use of this
material is permitted. Permission from IEEE must be obtained for all other uses,
in any current or future media, including reprinting/republishing this material
for advertising or promotional purposes, creating new collective works, for
resale or redistribution to servers or lists, or reuse of any copyrighted
component of this work in other works.}%

\maketitle

\begin{abstract}

Pruning seeks to design lightweight architectures by removing redundant weights
in overparameterized networks. Most of the existing techniques first remove
structured sub-networks (filters, channels,...) and then fine-tune the resulting
networks to maintain a high accuracy.  However, removing a whole structure is a
strong topological prior and recovering the accuracy, with fine-tuning, is
highly cumbersome. \\ 
In this paper,  we introduce an ``end-to-end'' lightweight network design that
achieves training and pruning simultaneously without fine-tuning. The design
principle of our method relies on reparametrization that learns not only the
weights but also the topological structure of the lightweight sub-network. This
reparametrization  acts as a prior (or regularizer) that defines pruning masks
implicitly from the weights of the underlying network, without increasing the
number of training parameters.  Sparsity is induced with a budget loss that
provides an accurate pruning. Extensive experiments conducted on the CIFAR10
 and the TinyImageNet datasets, using standard architectures (namely Conv4, VGG19 and ResNet18), show
compelling results without  fine-tuning.

\end{abstract}

%
\section{Introduction}
\label{sec:intro}

Deep neural networks (DNNs) have been highly effective  in solving many tasks
with several breakthroughs. However, the success of DNNs in different fields
comes at the expense of a significant increase of their computational overhead.
This dramatically limits their applicability, especially on cheap embedded
devices which are usually endowed with very limited computational resources.
Recent studies have shown that cumbersome yet powerful models are
overparametrized \cite{DBLP:conf/nips/DenilSDRF13} and can, therefore, be compressed to yield
compact and still effective models and different techniques have been introduced
in the literature in order to learn lightweight and highly efficient networks.

 \indent Existing  work tackles the issue of efficient network design either
 through knowledge distillation \cite{DBLP:journals/corr/HintonVD15} (and its  variants
 \cite{DBLP:conf/iclr/ZagoruykoK17,DBLP:journals/corr/RomeroBKCGB14,DBLP:conf/aaai/MirzadehFLLMG20,DBLP:conf/cvpr/ZhangXHL18,DBLP:conf/cvpr/AhnHDLD19}) or by tweaking
 network  parameters.  The latter is usually achieved using linear algebra
 \cite{DBLP:conf/nips/DentonZBLF14}, quantization,  binarization  as well as pruning
 \cite{DBLP:journals/corr/HanMD15,DBLP:journals/corr/abs-1902-05690,DBLP:conf/cvpr/JacobKCZTHAK18}. In particular, pruning aims at removing
 connections by zeroing the underlying weights while preserving high
 performances. Early work addresses unstructured pruning where the least
 important connections are removed
 individually~\cite{DBLP:conf/nips/CunDS89,DBLP:conf/nips/HassibiS92,DBLP:conf/nips/HanPTD15}. In most of the existing
 pruning methods, the key issue is to identify potentially irrelevant weights
 (possibly grouped) that could be removed. These candidates are usually
 identified with a saliency measure, and the most popular one is weight
 magnitude which seeks to  cancel weights with the smallest absolute values.

\indent Other more recent work focuses on structured
pruning~\cite{DBLP:conf/crv/RamakrishnanSN20,DBLP:conf/iclr/0022KDSG17,DBLP:journals/jetc/AnwarHS17} (specifically channel pruning
\cite{DBLP:conf/iccv/LiuLSHYZ17,DBLP:conf/icml/KangH20}) which allows significant and easily attainable speedup
factors on the current DNN platforms, at the expense of a coarser pruning rate.
However, structured pruning imposes a strong topological prior by removing whole
chunks in the primary networks, and achieves a lower sparsity rate compared to
unstructured pruning.  In order to mitigate the drop in accuracy that may occur
after pruning, existing methods rely on the evaluation of the Hessian matrix  of
the loss function which is intractable with the current massive neural
architectures  \cite{DBLP:conf/nips/CunDS89,DBLP:conf/icnn/HassibiSW93}. Other work (including Han et al.
\cite{DBLP:conf/nips/HanPTD15}) relies on an iterative pruning and fine-tuning, however, this
two-step process is highly cumbersome and requires several \emph{pruning and
fine-tuning} steps prior to converge to pruned networks whose performances
match, at some extent, the accuracy of the primary (unpruned) networks.

\indent In order to address the aforementioned issues, we introduce in this
paper a novel reparametrization that learns not only the weights of a surrogate
(lightweight) network but also its topology.  This reparametrization acts  as a
regularizer that models the tensor of the parameters of the surrogate network as
the Hadamard product of a weight tensor and an implicit mask. The latter makes
it possible to implement unstructured pruning constrained with a budget loss
that precisely controls the number of nonzero connections in the resulting
network. Experiments conducted on the CIFAR10 and the TinyImageNet classification tasks, using standard
primary architectures (namely Conv4, VGG19 and ResNet18), show the ability of
our method to train effective surrogate pruned networks  without any
fine-tuning.

\begin{figure*}
  \centerline{\includegraphics[width=12.5cm]{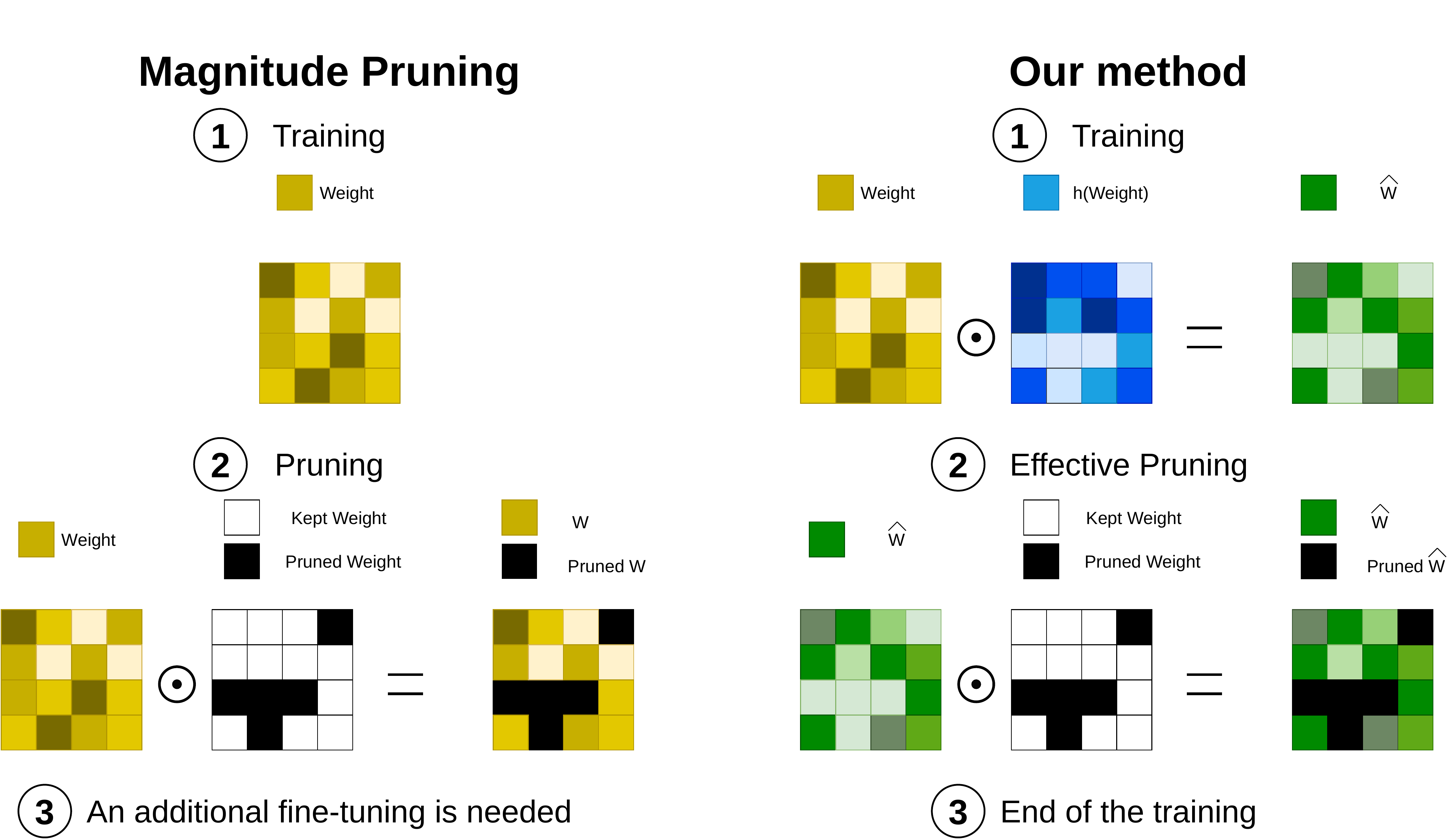}}
\caption{Comparison of our method and magnitude pruning. Magnitude pruning does
not include any prior on the weights during the initial training phase and needs
an additional fine-tuning procedure. Our method embeds a saliency heuristic
based on the weight magnitude in the weight reparametrization and does not
requires fine-tuning.}
\label{fig:comparison}
\end{figure*}

\section{Proposed Method}
\label{sec:method} 

Our reparametrization seeks to define a novel weight expression related to
magnitude pruning \cite{DBLP:conf/nips/CunDS89,DBLP:conf/nips/HanPTD15}.  This expression corresponds to
the Hadamard product involving a weight tensor and a function applied entry-wise
to the same tensor (as shown in \cref{fig:comparison}). This function acts as a
mask that  $i)$ multiplies weights by soft-pruning factors which capture their
importance and $ii)$ pushes less important weights to zero through a particular
budget  added to the cross entropy loss. \\
Our proposed framework allows for a joint optimization of the network weights and topology.  On the one hand, it prevents disconnections which may lead to degenerate networks with irrecoverable performance drop. On the other hand, it allows reaching a targeted pruning budget in a more 
convenient way than $\ell_1$ regularization.
Our reparametrization also helps minimizing the discrepancy between the primary
and the surrogate networks by maintaining  competitive performances without
fine-tuning. Learning the surrogate network requires only one step that achieves
pruning as a part of network design. This step  zeroes out the targeted number
of connections by constraining their reparametrized weights to vanish. 

\subsection{Weight reparametrization}

Without a loss of generality, we consider $f$ as a primary network whose layers
are recursively defined as
\begin{equation}
\label{eqn:layereq}
f(\mathbf{x}) = g_L \big(\mathbf{w}_L g_{L-1}(\mathbf{w}_{L-1}g_{L-2} \dots
\mathbf{w}_2 g_1(\mathbf{w}_1 \mathbf{x}))\big),
\end{equation}
\noindent with $g_\ell$ being a nonlinear activation associated to $\ell \in
\left\{ 1,\dots, L \right\}$ and $\left\{ \mathbf{w}_\ell \right\}_\ell$ a
weight tensor. Considering \cref{eqn:layereq}, a surrogate network $g$ is
defined as

\begin{equation}
\label{eqn:layereq2}
g(\mathbf{x}) = g_L \big(\mathbf{\hat w}_L g_{L-1}(\mathbf{\hat w}_{L-1}g_{L-2}
\dots\mathbf{\hat w}_2 g_1(\mathbf{\hat w}_1 \mathbf{x}))\big).
\end{equation}

\noindent In the above equation, $\mathbf{\hat w}_\ell$, referred to as
apparent weight, is a reparametrization of $\mathbf{w}_\ell$, that includes a
prior on its saliency, as 
\begin{equation}
  \label{eqn:reparam}
  \mathbf{\hat w}_\ell = \mathbf{w}_\ell  \odot h_t(\mathbf{w}_\ell),
\end{equation}
\noindent with $h_t$ being a function\footnote{with $t$ being its temperature
parameter.}  that enforces  the prior that smallest weights should be removed
from the network. In order to achieve this goal, $h_t$ should exhibit the
following properties:
\begin{enumerate}
  \item $\forall x \in \mathds{R},~~ 0 \leq h_t(x) \leq 1 $
  \item $h_t(x) \in C^1 \text{ on } \mathds{R}$
  \item $h_t(x) = h_t(-x)$
  \item $\forall a,\varepsilon \in\mathds{R}^{+\ast},~ \exists ~t
  \in\mathds{R}^{+\ast} ~ | ~ h_t(x) \leq \varepsilon, x \in [-a,a]$
\end{enumerate}

\begin{figure}[htb]

  \centering
  \centerline{\includegraphics[width=7cm]{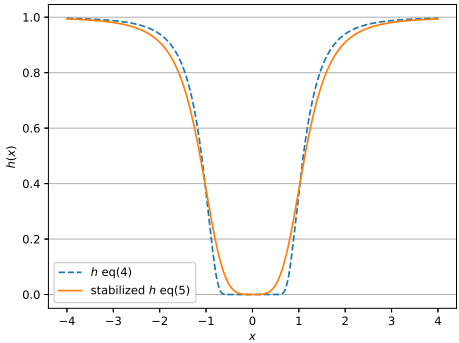}}
\caption{An example of a reparametrization function and its numerically stabilized variant, with $t=1$ and $n=4$.}
\label{fig:reparamfunctions}
\end{figure}

\noindent The first property ensures that the reparametrization  is neither
changing the sign of the apparent weight, nor acting as a scaling factor greater
than one. In other words, it  acts as the identity for sufficiently large
weights, and as a contraction factor for small ones. The second property is
necessary to ensure that the reparametrization function has computable gradient
while the third property states that $h_t$ should be symmetric  in order to
avoid  any bias towards the positive and the negative weights, so only their
magnitudes matter. The last property ensures the existence of  a temperature
parameter $t$  which allows upper-bounding the response of  $h_t$ on any
interval for any arbitrary $\varepsilon$. Hence,  $h_t$ acts as a stopband
filter which eliminates the smallest weights where  the parameter $t$ controls
the width of that filter.\\
In order to match a specific budget, the width of the stopband is tuned
according to the weight distribution of each layer. Note that the manual setting
of this parameter is difficult so in practice $t$ is learned as a part of
gradient descent; the initial setting  $t_\text{init}$ of this temperature is
shown in \cref{tbl:pruningperformances}.\\
Considering the aforementioned four properties of $h_t$, a simple choice of that
function is 
\begin{equation}
  \label{eqn:simpleh}
  h_t(x) = \exp\bigg\{{-\displaystyle\frac{1}{(tx)^n}}\bigg\}, ~ n\in 2\mathds{N},
\end{equation}
\noindent where $n$ controls the crispness of $h_t$. Although the function
described in \cref{eqn:simpleh} satisfies the four above properties, it suffers
from numerical instability as it generates  \texttt{NaN} (Not a Number) outputs
in most of the widely used  deep learning frameworks. We consider instead a
stabilized variant with a similar behavior, as \cref{eqn:simpleh},  that still
satisfies the four above properties (see also  \cref{fig:reparamfunctions}).
This numerically stable variant is  defined as 
\begin{equation}
  \label{eqn:realh}
  h_t(x) = C_1 \biggl( \text{exp} \bigg\{-\displaystyle\frac{1}{(tx)^n +1}\bigg\} - C_2 \biggr),
\end{equation}
\noindent with $C_1=\frac{1}{1-e^{-1}}$ and $C_2 = e^{-1}$; the constant added
to the denominator of the exponential  prevents numerical instability. Note that
the constants $C_1$ and $C_2$ are added to guarantee that $h_t$ satisfies  the
first property.

\subsection{Budget Loss}

Considering  $C(\{\mathbf{w}_1,\dots, \mathbf{w}_L\})$,  $C_\text{target}$ as
respectively the {\it actual} cost associated to a given network and the  {\it
targeted} one, the budget loss is defined as  
\begin{equation}
  \label{eqn:simplebudget}
  {\cal L}_\text{budget} = \bigl( C(\{\mathbf{w}_1,\dots, \mathbf{w}_L\}) - C_\text{target} \bigr)^2.
\end{equation} 

\noindent This budget loss is combined with the task classification loss. For a
better  conditioning of this combination, we normalize the budget loss by
$C_\text{initial}$; the latter corresponds to the cost of the primary unpruned
network and it is set in practice to the number of its parameters (see also
\cref{sec:experiments}). Hence, \cref{eqn:simplebudget} is updated as  
\begin{equation}
  \label{eqn:realbudget}
  {\cal L}_\text{budget} = \biggl( \displaystyle\frac{C(\{\mathbf{w}_1,\dots, \mathbf{w}_L\}) - C_\text{target}}{C_\text{initial}} \biggr)^2.
\end{equation}
\noindent Finally, the two losses are combined together via  a mixing parameter
$\lambda> 0$ that  controls the relative importance of  the budget, leading to
\begin{equation}
  \label{eqn:globalloss}
   {\cal L} =  {\cal L}_\text{task} + \lambda  {\cal L}_\text{budget}.
\end{equation}
\noindent Ideally, the budget of a neural network could be evaluated as the
FLOPs needed for a forward pass or through the $\ell_0$ norm of its weights.
However, neither is differentiable and therefore cannot be used in
a gradient-based optimization. In order to address this issue, we use our weight
reparametrization as a surrogate measure of $\ell_0$ and we define the cost
function as
\begin{equation}
  \label{eqn:ourcostfn}
  C(\{\mathbf{w}_1,\dots, \mathbf{w}_L\}) = \displaystyle \sum_{i=1}^{L} h(\mathbf{w}_i). 
\end{equation}

\section{Experiments}
\label{sec:experiments}

In this section, we evaluate the accuracy of our method using {Conv4}, {VGG19},
and {ResNet18} on the widely used  {CIFAR10} database and the {TinyImageNet}
database \cite{tinyimg}. The {Conv4} model is similar to the network used by Frankle et al. in
their Lottery Ticket experiments \cite{DBLP:conf/iclr/FrankleC19}, as a smaller
version of  {VGG19}  introduced by Simonyan et al.
\cite{DBLP:journals/corr/SimonyanZ14a} while  {ResNet18} is the small Residual
Network of  He et al. \cite{DBLP:conf/cvpr/HeZRS16}.  Comparisons w.r.t related
state of the art are shown on the  {CIFAR10} and {TinyImageNet} datasets.
{CIFAR10} is composed of $60.000$ color images of  $32 \times 32$ pixels,
divided into 10 classes. {TinyImageNet} is composed of $100.000$ color images of
$64 \time 64$ pixels divided into 200 classes. We use the default 2 fold setting for training and testing on both datasets.  

\subsection{Experimental setup}

Unless stated differently, all the networks have been trained during 300 epochs
with an initial learning rate of 0.1. A \emph{Reduce On Plateau} policy is
applied to the learning rate: if the test accuracy is not improving for 10
epochs in a row, then the learning rate is decreased  by a factor 0.3. A weight
decay is applied on the weights with a penalization factor of $5 \times
10^{-5}$. An \emph{Early Stopping} policy was used to stop prematurely the
training if no improvement of the test accuracy is observed  in 60 epochs. This
procedure has been used both for initial training, and for fine-tuning of
magnitude pruning (MP). \\ Performances are reported on the pruned networks: the
latters are effectively pruned (extremely small weights are set to zero) up to
the targeted pruning rate.

\begin{table}[htbp]
  \centering\begin{tabular}{|c||c|c|c|c|c|}
  \hline
\textbf{Network}                 & \textbf{Method} & \textbf{90\%}    & \textbf{95\%}    & \textbf{97\%}    & \textbf{99\%}    \\
  \hline
   \multicolumn{6}{|c|}{Dataset: \textbf{CIFAR10}}  \\
  \hline 
\multirow{3}{*}{Conv4}    & Mag    & 54.94 & 22.91 & 13.1  & 13.07 \\
  \cline{2-6}
                          & Mag+FT & 89.50 & 87.38 & 85.57 & 76.7  \\
  \cline{2-6}
                          & Ours   & 85.57 & 84.59 & 82.6  & 16.54 \\
  \hline
  \cline{2-6}
\multirow{3}{*}{VGG19}    & Mag    & 10.0  & 10.0  & 10.0  & 10.0  \\
  \cline{2-6}
                          & Mag+FT & 93.57 & 93.52 & 92.96 & 10.0  \\
  \cline{2-6}
                          & Ours   & 86.03 & 47.08 & 10.0  & 10.0  \\
  \hline
  \cline{2-6}
\multirow{3}{*}{ResNet18} & Mag    & 77.89 & 37.55 & 14.19 & 10.0  \\
  \cline{2-6}
                          & Mag+FT & 94.69 & 94.43 & 94.03 & 92.25 \\
  \cline{2-6}
                          & Ours   & 91.59 & 89.89 & 33.49 & 15.08 \\
  \hline
  \multicolumn{6}{|c|}{Dataset: \textbf{TinyImageNet}}  \\
  \hline
  \multirow{3}{*}{Conv4}    & Mag    & 26 & 10 & 3  & 0.5 \\
    \cline{2-6}
                            & Mag+FT & 45 & 45 & 43 & 21  \\
    \cline{2-6}
                            & Ours   & 39 & 39 & 39  & 35 \\
    \hline
  
  \end{tabular}
  \caption{Performances (accuracy) on the \textbf{CIFAR10} and \textbf{TinyImageNet} test datasets for
  magnitude pruning (with and without fine-tuning) and our method, for different
  target pruning rates $p$. Our results are shown for $\lambda=5$, $n=4$, $t_\text{init}=100$ and $C_\text{target}=p \times C_\text{initial}$}
  \label{tbl:pruningperformances}
\end{table}

\subsection{Pruning performances}

Table\ref{tbl:pruningperformances} shows a comparison of our method against MP
with and without fine-tuning. The performance is reported as the accuracy
obtained on the test set of {CIFAR10}. For small networks such as {Conv4}, our
method achieves compelling performances up to high pruning rates ($97$\%)
without any fine-tuning. The performances are within a few percent of fine-tuned
MP (less than $4$\% up to a pruning rate of $97$\%), while providing a
significant  improvement w.r.t. the non fine-tuned version of MP (at least more
than $30$\% for pruning rates up to $97$\%). For larger networks (namely {VGG19}
and  {ResNet18}), the significant improvement over the non fine-tuned version of
MP is still observed, although less pronounced. For high pruning rates, our
method lags behind the fine-tuned version as shown in these results.

Observed  performances   in \cref{tbl:pruningperformances} can be explained by
analyzing the {\it actual} fraction of the remaining weights for the different
{\it targeted} pruning rates shown in \cref{tbl:realizedpruningrate}. This
fraction is obtained by dividing the actual cost of the network (once its
training achieved) by the total number of its parameters. This cost is computed
with our function defined in \cref{eqn:ourcostfn}. Table
\ref{tbl:realizedpruningrate} shows that the targeted pruning is not always
precisely reached, specifically for the most extreme (high) rates. When the
targeted rate is not precisely reached, this has a strong negative impact on the
effectively pruned  networks. In contrast, when the targeted rate is reached,
the test accuracy is either within a small margin (less than $0.5$\%), or larger
than the performances of the network trained with our method, before effective
pruning. This is a significant improvement w.r.t. MP, which results into a
dramatic drop in performance if fine-tuning is not applied. Results also show
that our method successfully trains the weights while preparing the network for
sparsity by reducing the magnitude of the less important weights. The
performances of the networks trained with our method before effective pruning
are reported in \cref{tbl:perfbeforepruning} for the different targeted rates.

\noindent Finally, on TinyImageNet, our method remains stable w.r.t. different pruning rates and outperforms MP as well as fine-tuned MP by a significant margin, especially at the highest pruning rates.

\begin{table}[]
  \centering\begin{tabular}{|l||l|l|l|l|}
    \hline
     \textbf{Network} & \textbf{90}\%  & \textbf{95\%}  & \textbf{97\%}   &
     \textbf{99\%}   \\
    \hline
  Conv4    & 0.10 & 0.05 & 0.03 & \emph{0.03} \\
    \hline
  VGG19    & \emph{0.11} & \emph{0.09} & \emph{0.08} & \emph{0.08} \\
    \hline
  ResNet18 & 0.09 & 0.05 & \emph{0.06} & \emph{0.05} \\
    \hline
  \end{tabular}
  \caption{Fraction of the remaining weights for targeted pruning rates on the
  \textbf{CIFAR10} dataset, with $\lambda=5$, $n=4$ and $t_\text{init}=100$,
  computed with our surrogate cost function (cf \cref{eqn:ourcostfn}).
  \emph{Italics} design fractions of the remaining weights that are in excess
  compared to the targeted pruning rate.}
  \label{tbl:realizedpruningrate}
\end{table}

\begin{table}
  \centering\begin{tabular}{|l||l|l|l|l|}
    \hline
     \textbf{Network} & \textbf{90}\%  & \textbf{95\%}  & \textbf{97\%}   &
     \textbf{99\%}   \\
    \hline
  Conv4    & 85.68 & 83.04 & 84.89 & 82.68 \\
    \hline
  VGG19    & 88.65 & 88.07 & 88.89 & 89.19  \\
    \hline
  ResNet18 & 91.62 & 91.01 & 90.42 & 90.71 \\
    \hline
  \end{tabular}
  \caption{Performances (accuracy) on the \textbf{CIFAR10} test dataset before
  effective pruning, for different target pruning rates, with $\lambda=5$, $n=4$
  and $t_\text{init}=100$}
  \label{tbl:perfbeforepruning}
\end{table}

\section{Conclusion}
\label{sec:conclusion}

We introduce in this paper a new budget-aware pruning method based on weight
reparametrization. The latter acts as a regularizer that emphasizes on the most
important connections in a given primary network while reaching a targeted cost
and  maintaining relatively close performances. Extensive experiments conducted
on standard networks and datasets show that our method achieves compelling
results without any fine-tuning. Moreover, while the targeted budget is
satisfied, our method does not show any significant performance degradation even
when enforced with an effective pruning step. Future work will investigate
other classification tasks requiring surrogate networks with higher pruning
rates and close performances w.r.t. the underlying primary networks. In
particular online training on massive data (such as videos) is time demanding,
and requires training cost-aware lightweight networks very efficiently. \\ 

\noindent \noindent\textbf{Acknowledgment.} This work has been achieved within a partnership between Sorbonne University (LIP6 Lab) and Netatmo.



\vfill\pagebreak
\bibliographystyle{IEEEbib}
\bibliography{dupont}

\end{document}